\documentclass[letterpaper, 10 pt, conference]{ieeeconf}  

\IEEEoverridecommandlockouts                              

\usepackage{graphics} 
\usepackage{epsfig} 
\usepackage{mathptmx} 
\usepackage{times} 
\usepackage{amsmath} 
\usepackage{amssymb}  
\usepackage{booktabs}
\usepackage{url}
\usepackage{subcaption}
\usepackage{graphicx}  
\usepackage[hidelinks]{hyperref}

\title{\LARGE \bf
Evo-0: Vision-Language-Action Model with Implicit Spatial Understanding
}

\author{Tao Lin$^{*}$,\; Gen Li$^{*}$,\; Yilei Zhong,\; Yanwen Zou,\; Yuxin Du,\; Jiting Liu,\; Encheng Gu$^{4}$,\; Bo Zhao$^\dagger$\\
$^{1}$School of AI, Shanghai Jiao Tong University,\; $^{2}$EvoMind Tech,\; $^{3}$IAAR-Shanghai,\; $^{4}$University of Cambridge\;\\
\texttt{taolin200108@gmail.com, bo.zhao@sjtu.edu.cn} \\
\url{https://mint-sjtu.github.io/Evo-0.io/}
}

\makeatletter
\long\def\@makecaption#1#2{%
  \vskip 10pt%
  \footnotesize
  \centering
  \textbf{#1.} #2 
  \vskip 5pt%
}
\makeatother

\begin{document}

\maketitle
\thispagestyle{empty}
\pagestyle{empty}

\begin{abstract}

Vision-Language-Action (VLA) models have emerged as a promising framework for enabling generalist robots capable of perceiving, reasoning, and acting in the real world. These models usually build upon pretrained Vision-Language Models (VLMs), which excel at semantic understanding due to large-scale image and text pretraining. However, existing VLMs typically lack precise spatial understanding capabilities, as they are primarily tuned on 2D image-text pairs without 3D supervision. 
To address this limitation, recent approaches have incorporated explicit 3D inputs such as point clouds or depth maps, but this necessitates additional depth sensors or pre-trained depth estimation models, which may yield defective results.
In contrast, our work introduces a plug-and-play module that implicitly incorporates 3D geometry features into VLA models by leveraging an off-the-shelf visual geometry foundation model.
This integration provides the model with depth-aware visual representations, improving its ability to understand the geometric structure of the scene and the spatial relationships among objects from RGB images alone. 
We evaluate our method on a set of spatially challenging tasks in both simulation and the real world. Extensive evaluations show that our method significantly improves the performance of state-of-the-art VLA models across diverse scenarios.

\end{abstract}

\section{Introduction}

Vision-Language-Action (VLA) models have recently attracted substantial attention and achieved notable progress. 
These models typically fine-tune pre-trained Vision-Language Models (VLMs) using robot manipulation data to leverage the vision-language generalization ability learned from large-scale image-text data.
This paradigm has achieved impressive results across a wide range of real-world and simulated tasks.

However, existing VLA models exhibit a critical limitation: the lack of precise 3D spatial understanding.
This limitation can be largely attributed to two main factors: 
(1) the pre-training data and objectives of VLMs are primarily based on 2D image-text alignments, and
(2) the robotic datasets used for fine-tuning typically contain only RGB observations, lacking 3D spatial information.
As a result, these models often struggle to capture the precise geometric and spatial relationships that are essential for effective interaction in the physical world.
Recent studies~\cite{spatialbot,mm-spatial} have empirically validated this observation, showing that VLMs tend to generalize poorly when it comes to interpret 3D structures from visual inputs alone. This presents a critical bottleneck in scaling VLA models to more complex tasks that require accurate spatial reasoning and physical interaction.

To address this limitation, recent approaches~\cite{pointvla,spatialvla,3d-vla,bridgevla} have tried to incorporate 3D information into VLA models to enhance their spatial understanding capabilities. A common strategy is to explicitly inject depth information into the learning pipeline, such as point clouds or depth maps, either captured by depth sensors or predicted by pre-trained depth estimation networks~\cite{spatialbot,bhat2023zoedepth,yang2024depth}.
While effective to some extent, these methods introduce new challenges. They often require additional depth sensors, which may not be available in practical settings. Moreover, the defective depth estimations can introduce extra noise, affecting the reliability of the learned 3D representations.

In this paper, we introduce \emph{Evo-0}, a novel VLA architecture that explores an implicit strategy to enhance the spatial understanding of VLA models without relying on depth sensors or explicit depth estimation. We leverage a spatial encoder trained on large-scale 2D–3D paired data \cite{vggt} to extract geometric features from RGB images, thereby enhancing the spatial understanding of VLA.
To this end, we design a lightweight fusion module that integrates geometry features from spatial encoder with visual tokens in VLM, enabling the model to perceive object layouts and reason about spatial relations more effectively. 
We demonstrate the effectiveness of our method through comprehensive experiments on 5 simulation tasks, 5 real-world manipulation tasks, and a robustness evaluation with 5 distinct disturbance conditions. Across all settings, our model consistently enhances spatial understanding and outperforms state-of-the-art VLA models.

Our contributions are listed as follows: (1) We propose a plug-and-play module to enhance the spatial understanding of VLA models by implicitly injecting 3D geometric priors without using depth sensors or explicit depth estimation. (2) We evaluate our method on a diverse set of spatially challenging tasks in both simulation and the real world, demonstrating consistent improvements over strong baselines. (3) We design a robustness evaluation setup under multiple disturbance conditions to validate our method’s effectiveness in real-world perturbations.

\section{Related Work}

\noindent \textbf{Vision-Language-Action Models.}
Recently, several studies \cite{kim2024openvla,black2410pi0,bjorck2025gr00t,bridgevla,liu2024rdt,brohan2023rt} have focused on building general-purpose robot policies by extending pre-trained vision-language models (VLMs) with action prediction capabilities. These models, known as vision-language-action (VLA) models, demonstrate strong performance and few-shot generalization across a wide range of embodied tasks.

Among them, OpenVLA~\cite{kim2024openvla} is trained on 970k multi-robot demonstrations from the Open-X Embodiment~\cite{o2024open} dataset, demonstrates strong generalization across a wide range of tasks and embodiments, and supports efficient fine-tuning under limited computational resources. 
$\pi_0$~\cite{black2410pi0} adapts the PaliGemma~\cite{beyer2024paligemma} architecture for robotic control and introduces a flow-matching-based~\cite{lipman2022flow,liu2022rectified} action expert module that enables accurate prediction of continuous actions. GR00T~\cite{bjorck2025gr00t} introduces an effective co-training strategy that jointly leverages web data, synthetic data, and real-world robot data within a unified framework, enabling broad generalization across tasks and embodiments.

Despite the promising progress, most existing VLA models primarily rely on 2D visual inputs and lack effective mechanisms for modeling the 3D spatial structure of the scene, which limits their spatial reasoning capabilities in complex manipulation tasks.

\noindent \textbf{Robot Learning with 3D Information.}
In response to the spatial limitations of 2D-based VLA models, several recent approaches~\cite{spatialbot, 3d-vla,pointvla,spatialvla,chen2024sugar,goyal2024rvt,jia2024lift3d} have explored integrating 3D information to enhance spatial understanding. 
For example, 3D-VLA~\cite{3d-vla} fuses 3D perception, reasoning, and action through a 3D-based large language model~\cite{3d-llm}, trained on a large-scale 3D dataset curated from existing embodied robotics benchmarks.
To make 3D-aware policies applicable in real-world scenarios, methods such as SpatialVLA~\cite{spatialvla} and PointVLA~\cite{pointvla} incorporate depth information captured from additional RGB-D cameras or depth estimation models, which enhances 3D scene understanding and enables more accurate perception of spatial relationships, object geometry, and depth-aware interactions.

Despite these advances, a fundamental limitation of current 3D-aware VLA methods lies in their reliance on explicit 3D inputs such as depth maps and point clouds, which require either specialized sensors or auxiliary estimation models. This dependency imposes constraints on scalability, deployment flexibility, and general applicability in diverse real-world environments.

To address this issue, we propose integrating VGGT~\cite{vggt} into existing VLA models. While keeping the input as RGB images, VGGT implicitly models 3D structure by fusing spatial features from multi-view observations. Our approach serves as a bridge between pure 2D input models and explicit 3D perception methods, enhancing spatial understanding without requiring additional sensors or depth estimation modules.
\begin{figure*}[!t]
  \centering
  \includegraphics[width=\textwidth]{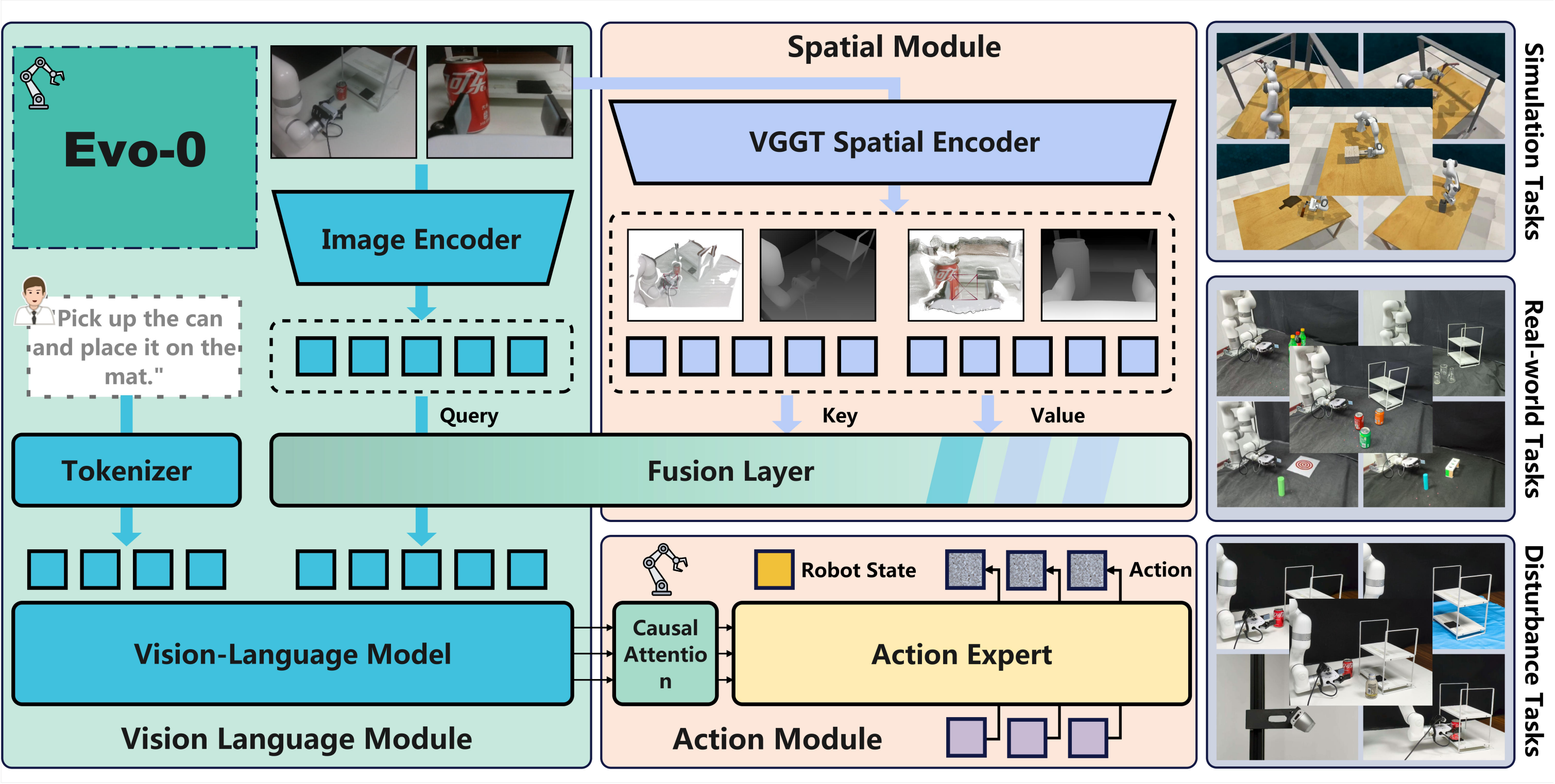} 
\captionsetup{justification=centering}
  \caption{\textbf{Architecture of Evo-0.} The input RGB images are initially processed by a 2D image encoder and a VGGT spatial encoder. The extracted features are subsequently fused through a fusion layer to form a spatially enriched visual representation. This representation is further propagated through a vision language model and an action module to produce the robot actions. This spatially-enhanced perception pipeline leads to strong performance across the simulation, real-world, and disturbance tasks illustrated on the right side of the figure.}
  \label{fig:archi}
\end{figure*}
\section{Method}


\subsection{Preliminaries}
\noindent \textbf{Vision-Language-Action Models.}  
As a promising approach toward generalist robot policies, Vision-Language-Action (VLA) models have emerged as an increasingly popular research direction~\cite{black2410pi0,kim2024openvla,pointvla,liu2025hybridvla}. VLAs aim to bridge the gap between high-level human instructions and low-level robotic actions by leveraging the rich multimodal priors encoded in large-scale pre-trained Vision-Language Models (VLMs), such as Paligemma~\cite{beyer2024paligemma}, CLIP~\cite{radford2021learningtransferablevisualmodels}, LLaMA~\cite{llama,llama2}, and Flamingo~\cite{alayrac2022flamingovisuallanguagemodel}. These VLMs are trained on vast and diverse internet-scale image-text pairs, endowing them with strong world knowledge and the ability to ground natural language in visual concepts.

Unlike traditional imitation learning methods that typically train a task-specific policy from scratch, VLA models reuse this pretrained multimodal understanding to enable more flexible and scalable robotic behaviors. In particular, the VLM serves as a general-purpose semantic encoder, while a downstream module---commonly referred to as the \textit{action expert}---learns to map the fused representations into robot control commands. This modular design separates general world understanding from task-specific actuation, allowing the model to generalize better across instructions and visual environments.

Formally, at each timestep $t$, the VLA model receives multi-view visual observations $\{I_t^i\}_{i=1}^N$ and a language instruction $L$, which are jointly encoded by the VLM to produce a contextual embedding $z_t$. This embedding is then concatenated with robot-specific states $S_t$ (e.g., joint angles, gripper status, or end-effector pose), and passed to the action expert to generate the low-level control command $A_t$. The entire pipeline thus defines a conditional distribution $p(A_t \mid I_t^i, L, S_t)$.

Compared to standard imitation learning policies, which are typically trained on a specific task, the VLA framework improves \textit{semantic grounding}, \textit{modality fusion}, and \textit{generalization capability}. This enables robots not only to follow diverse and abstract language instructions but also to adapt to new tasks and visual scenes with minimal fine-tuning.

\noindent \textbf{Visual Geometry Foundation Models.}
Unlike traditional SLAM or depth estimation pipelines that rely on finely-tuned modules and sensors, Visual Geometry Foundation Models (VGFMs)~\cite{mast3r,Dust3r,vggt,megasam} are a class of vision models trained to reconstruct 3D structural information from 2D visual inputs. Since VGFMs are trained with geometric supervision, they have the ability to recover fine-grained spatial structure from multi-view monocular inputs. These models provide strong structural priors for downstream tasks such as spatial understanding, especially when explicit 3D sensors are unavailable.

Given a set of multi-view images $\{I^i\}_{i=1}^N$, a typical VGFM predicts a 3D point cloud $P$ representing the scene as 
\begin{equation}
f_{\text{VGFM}}(\{I^i\}_{i=1}^N)=P.    
\end{equation}
These geometry-aware models complement vision-language systems by injecting 3D structural cues, enhancing spatial grounding from purely 2D observations such as video frames.

Recently, Visual Geometry Grounded Transformer (VGGT)~\cite{vggt} has introduced a novel feed-forward architecture and demonstrated impressive performance in 3D attributes prediction.
It takes an arbitrary number of image views as input and alternates between frame-wise and global self-attention to model spatial consistency. 
Given a sequence of $N$ RGB images $\{I^i\}_{i=1}^N$, where each $I_i \in \mathbb{R}^{3 \times H \times W}$, the model outputs a set of 3D annotations for each frame, including predicted camera poses $g_i$, depth maps $D_i$, point maps $P_i$, and 3D point tracks $T_i$, i.e.,
\begin{equation}
f\left(\{I_i\}_{i=1}^N\right) = \left(g_i, D_i, P_i, T_i\right)_{i=1}^N.
\end{equation}

\subsection{Proposed VLA Architecture}

\begin{figure*}[!t]
  \centering
  \includegraphics[width=\linewidth]{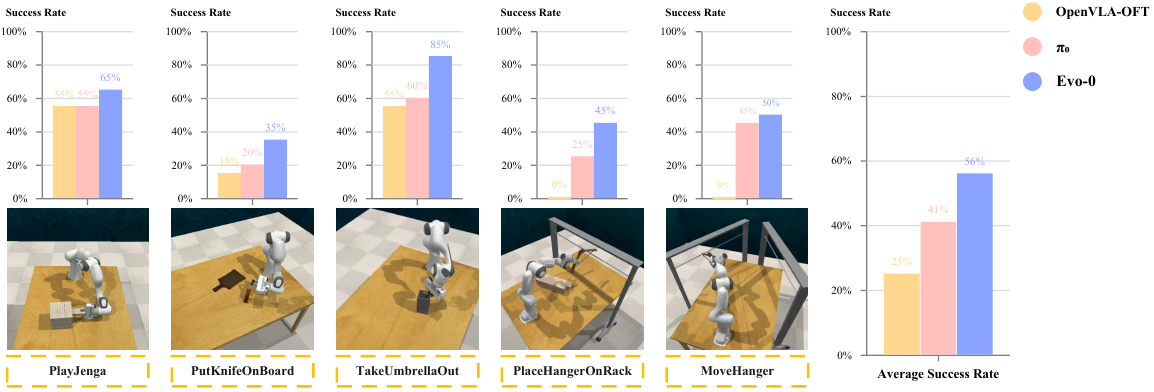}
  \caption{\textbf{Simulation Experiments.} We evaluate all models on five RLBench simulation tasks requiring precise spatial manipulation, using a multi-task training setup. Success rates are shown per task and averaged. The success criteria are defined according to the official RLBench specifications.
}
  \label{fig:Simulation_experiments}
\end{figure*}

Recent 3D-based VLA models, such as PointVLA~\cite{pointvla} and SpatialVLA~\cite{spatialvla}, often employ explicit 3D inputs like point clouds or depth maps to enhance spatial understanding.
While effective, these approaches typically require additional sensors and preprocessing, and are often sensitive to variations in camera viewpoints.
In contrast, VGGT presents a promising alternative for implicitly introducing spatial awareness, benefiting from its diverse training data and elegant feed-forward architecture.
Recent studies have successfully applied VGGT to VLM architectures~\cite{wu2025spatialmllmboostingmllmcapabilities} and SLAM systems~\cite{maggio2025vggtslamdensergbslam}, demonstrating that geometry-grounded visual tokens can improve spatial understanding in both multimodal learning and classical robotic perception.

Motivated by these findings, we hypothesize that introducing geometry-aware visual representations from VGGT into the action prediction pipeline can enrich spatial context, leading to more precise and generalizable policy learning without requiring explicit point cloud or depth inputs.
To evaluate this hypothesis, we build our model upon $\pi_0$~\cite{black2410pi0}, a state-of-the-art open-source VLA model, and incorporate geometry-aware features from VGGT into its visual embedding stream. The architecture is described in Figure~\ref{fig:archi}.
Specifically, we utilize VGGT as a spatial encoder and extract tokens from its final layer:
\begin{equation}
    \mathcal{E}_{l}\left(\{I_i\}_{i=1}^N\right) = t_{c}, t_{r}, t_{3D},
\end{equation}
where $N$ is the number of views, $l$ denotes the layer index, and $t_c$, $t_r$, and $t_{3D}$ denote the camera, register, and 3D tokens, respectively.
We extract the 3D tokens $t_{3D}$ to inject spatial information, as they are originally trained to conduct 3D tasks in VGGT.
These tokens capture rich geometric representations, including depth-aware context, temporally consistent object trajectories, and spatial correspondences across views.

To integrate the VGGT-derived token features into the vision-language pipeline, we introduce a lightweight fuser module that combines embeddings from the Vision Transformer~\cite{dosovitskiy2020image} and the VGGT encoder. 
Specifically, the fuser consists of a single cross-attention layer, where the 2D visual tokens $t_{2D} \in \mathbb{R}^{N \times M_{2D} \times d_{2D}}$ serve as queries, and the VGGT-derived tokens $t_{3D} \in \mathbb{R}^{N \times M_{3D} \times d_{3D}}$ act as keys and values. Here, $M_{2D}$ and $M_{3D}$ denote the number of tokens from ViT and VGGT encoder, respectively. 
The 2D visual tokens are then updated as follows:
\begin{equation}
Q = t_{2D} W_Q, \quad K = t_{3D} W_K, \quad V = t_{3D} W_V,
\end{equation}
\begin{equation}
t^{i} = \text{softmax}\left( \frac{Q^{i} (K^{i})^\top}{\sqrt{d}} \right) V^{i},
\end{equation}
\begin{equation}
t = \text{Concat}_{i=1}^{N} \left( t^{i} \right),
\end{equation}

where $W_Q\in \mathbb{R}^{d_{2D} \times d}$, and $W_K$, $W_V$ $\in \mathbb{R}^{d_{3D} \times d}$ are trainable projection matrices shared across views. 
Each view $i\in {1,...,N}$ is processed independently via the cross-attention module, and the resulting tokens are concatenated to form the fused output $t$.

The fused tokens are then forwarded to the PaliGemma~\cite{beyer2024paligemma} vision-language model, which jointly attends over both the geometry-enhanced visual input and the language tokens to predict actions. 
To maintain computational efficiency and minimize disruption to the pretrained VLM backbone, we freeze the core VLM parameters and insert lightweight Low-Rank Adaptation (LoRA)~\cite{lora} layers. 
During training, only the fuser module, LoRA layers, and the flow-matching action expert are fine-tuned, enabling effective adaptation with minimal overhead.

\section{Experiments}
This section presents a series of experiments conducted in both simulation and real-world settings to evaluate the spatial perception capabilities of our model. We show that Evo-0 consistently outperforms state-of-the-art baselines across a range of manipulation tasks. In Section A, we compare Evo-0 with competitive baselines in the RLBench simulation environment, focusing on fine-grained object grasping and transport. In Section B, we analyze the influence of key hyperparameters on manipulation accuracy. In Section C, we assess the model’s ability to perform precise manipulations in real-world environments. In Section D, we evaluate the robustness of Evo-0 under unseen disturbances, including novel distractor objects, background changes, spatial displacements, height variations, and changes in camera viewpoint.

\begin{figure*}[!t]
  \centering
  \includegraphics[width=\linewidth]{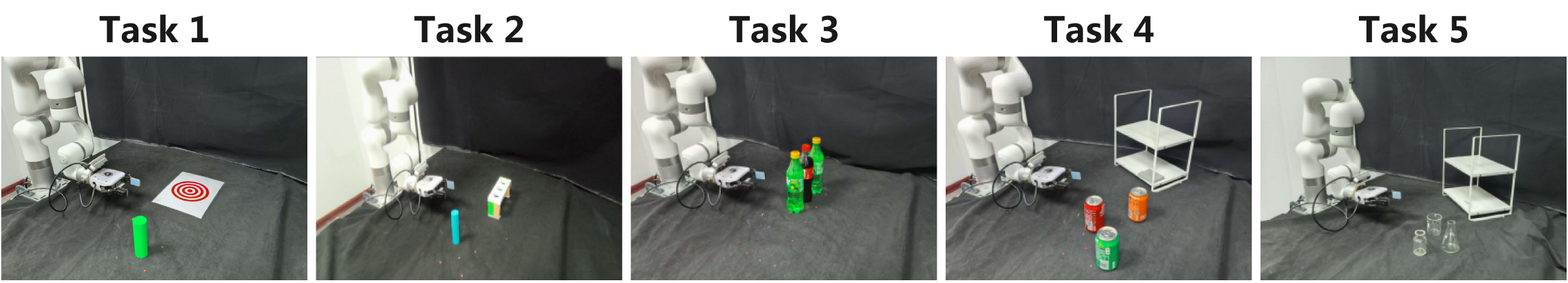}
  \caption{\textbf{Illustration of the task setup.} Five real-world evaluation tasks are used in our experiments: centering a cylinder on a target, peg-in-hole insertion, middle bottle grasping, can pick-and-place, and transparent object pick-and-place.}
  \label{fig:task_setup}
\end{figure*}

\subsection{Simulation Experiments}
    \noindent \textbf{Task Setup.} To enable systematic evaluation, we conduct experiments using the RLBench benchmark within the CoppeliaSim simulator. We select five tasks that require precise spatial reasoning and fine-grained manipulation to construct our evaluation suite. Each task is executed using a Franka Panda robot equipped with a front camera, a wrist camera, and an overhead camera. The selected tasks include \textit{PlayJenga}, \textit{PutKnifeOnChoppingBoard}, \textit{TakeUmbrellaOutOfUmbrellaStand}, \textit{PlaceHangerOnRack} and \textit{MoveHanger}. The selected tasks cover three categories of precise manipulation which demand strong spatial understanding:
\begin{enumerate}
    \item \textbf{precise grasping and transport.} (e.g., \textit{PlayJenga}, \textit{TakeUmbrellaOutOfUmbrellaStand})
    \item \textbf{precise grasping and placement.} (e.g., \textit{PutKnifeOnChoppingBoard})
    \item \textbf{precise motion under height and translation variation.} (e.g., \textit{PlaceHangerOnRack}, \textit{MoveHanger})
\end{enumerate} We construct the training dataset using the official data generation scripts provided by the RLBench repository, collecting 100 demonstration trajectories for each task.

\noindent \textbf{Implementation Details.} Our framework builds upon the open-source VLA model \cite{black2410pi0}. We adopt OpenVLA-OFT \cite{kim2025fine}  and $\pi_0$ as the baseline for comparison. The input comprises three RGB images, each resized to 224×224. Both the input state and the predicted output are represented as absolute joint angles. For a fair comparison, we utilize the official pretrained parameters provided by each method and adhere to their original training settings. For Evo-0, model is trained using the AdamW optimizer with a weight decay of $10^{-10}$ and a cosine learning rate schedule: the learning rate starts at $2.5 \times 10^{-5}$, warms up over 1000 steps, and decays to $2.5 \times 10^{-6}$. Training is conducted on a single NVIDIA A800 GPU (80 GB) with bfloat16 mixed precision, using a batch size of 32. For evaluation, we perform 20 rollouts for each task to assess success rates.

\noindent \textbf{Quantitative Results.}
As shown in Figure~\ref{fig:Simulation_experiments}, Evo-0 consistently outperforms OpenvVLA-OFT and $\pi_0$ across all five tasks.  
Specifically, OpenVLA-OFT achieves success rates of 55\%, 15\%, 55\%, 0\% and 0\% on \textit{PlayJenga}, \textit{PutKnifeOnChoppingBoard}, \textit{TakeUmbrellaOutOfUmbrellaStand}, \textit{PlaceHangerOnRack} and \textit{MoveHanger}, respectively, resulting in an average of 25\%. Similarly, $\pi_0$ obtains 55\%, 20\%, 60\%, 25\% and 45\% on the same tasks, yielding an average of 41\%.
In contrast, Evo-0 attains success rates of 65\%, 35\%, 85\%, 45\% and 50\% on the same tasks, with an average of 56\%---reflecting an improvement of 31 percentage points over OpenVLA-OFT and 15 percentage points over $\pi_0$, respectively.

Among all tasks, the most notable improvements over $\pi_0$ are observed on \textit{TakeUmbrellaOutOfUmbrellaStand} and \textit{PlaceHangerOnRack}, with gains of 25 and 20 percentage points, respectively.
 These results indicate that Evo-0 demonstrates superior spatial perception and generalization capabilities, particularly in scenarios that involve precise localization, trajectory planning and manipulation of collision-sensitive objects. This also validates the effectiveness of our proposed spatial fusion module, which substantially enhances the model’s spatial perception and reasoning capabilities.

\noindent \textbf{Inference Speed.}
We evaluate the control frequency of $\pi_0$ and Evo-0 on NVIDIA RTX 4090 GPU. The control frequency of $\pi_0$ reaches 11.3~Hz, while that of Evo-0 is reduced to 6.94~Hz, primarily due to the increased computational cost of image encoding via the VGGT module. Despite this overhead, Evo-0 maintains a frequency well within the range required for real-time robotic control.

\begin{figure}[t]
  \centering
  \begin{subfigure}[b]{0.48\linewidth}
    \centering
    \includegraphics[width=\linewidth]{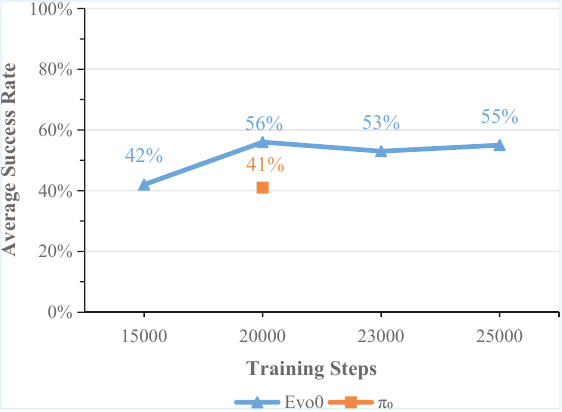}
    \caption{}
    \label{fig:a}
  \end{subfigure}
  \hfill
  \begin{subfigure}[b]{0.48\linewidth}
    \centering
    \includegraphics[width=\linewidth]{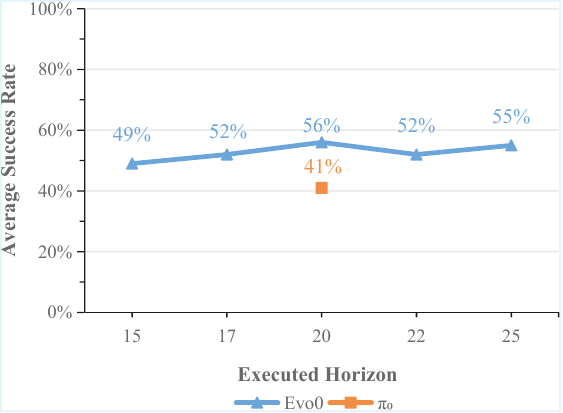}
    \caption{}
    \label{fig:b}
  \end{subfigure}
  \caption{\textbf{Hyperparameter Experiments.} We visualize the impact of training steps and horizon on success rates. The results highlight the training efficiency of our method and its robustness to horizon variation.}
  \label{fig:hyper_results}
\end{figure}

\subsection{Hyperparameter Experiments}
To analyze how hyperparameters affect model performance, we conduct additional experiments on the five RLBench tasks. We focus on two aspects: the number of training steps and the executed horizon, and evaluate their influence on task success rates.

\noindent \textbf{Training Steps.}
To examine how training progress affects policy performance, we evaluate Evo-0 using checkpoints saved at 15k, 20k, 23k, and 25k training steps. As illustrated in Figure~\ref{fig:hyper_results}(a), the success rate consistently improves and gradually converges as training proceeds. Notably, Evo-0 trained with only 15k steps already outperforms $\pi_0$ trained with 20k steps, highlighting the superior training efficiency of our method.

\noindent \textbf{Executed Horizon.}
To analyze the impact of the executed horizon on policy performance, we evaluate Evo-0 (trained with 20k steps) under five different horizon settings: 15, 17, 20, 22 and 25. As shown in Figure~\ref{fig:hyper_results}(b), Evo-0 maintains stable performance across all configurations, achieving average success rates of 49\%, 52\%, 56\%, 52\% and 55\%, respectively. It consistently outperforms $\pi_0$ (41\% average) across all horizons, demonstrating robustness to variations in execution length. Although minor fluctuations are observed across tasks, no setting leads to a significant performance degradation, indicating that Evo-0 is largely insensitive to the executed horizon and preserves reliable spatial reasoning and trajectory control.

\subsection{Real-World Experiments}

\noindent \textbf{Task Setup.}
We design five tasks for real-world robot evaluation, which span a range of spatial understanding challenges, from fine-grained geometric alignment to pick-and-place and transparent object interaction.
In particular, each task has a low tolerance for spatial error, as minor inaccuracies in the spatial predictions can lead to task failure.
This makes them well-suited for assessing if representations from VGGT can enhance VLA's spatial understanding.

A detailed description of the five tasks is provided below, with an illustration shown in Figure~\ref{fig:task_setup}.
\begin{enumerate}
    \item \textbf{Centering a cylinder on a target.} The robot is required to align a cylindrical object precisely at the center of a marked target area on the table. This task resembles target shooting: the target has concentric rings, and scoring is based on which ring the center of the cylinder falls into. The closer to the center, the higher the score. 
    \item \textbf{Peg-in-hole insertion.} This task requires the robot to insert a cylindrical peg into one of three tightly fitting holes on a board. This necessitates accurate alignment in 3D space, as small tilting or offset could cause task failure.
    \item \textbf{Middle bottle grasping.} Three bottles are closely placed in a row, and the robot is instructed to pick the middle one. This setup mimics a grocery store scenario, where items are densely arranged on shelves. Success is defined as picking up the middle bottle without knocking over the adjacent ones.
    \item \textbf{Can pick-and-place.} In this task, the robot must pick up a standard can and place it in a designated spot on a shelf. The location of the placement is varied across trials in both position and height, requiring the model to generalize spatial understanding to different configurations.
    \item \textbf{Transparent object pick-and-place.} The task setup is similar to the previous one, but involves transparent objects such as glass bottles.
    This presents additional challenge, since transparent materials are often poorly captured by RGB sensors and are prone to glare, making them difficult to perceive and localize.

\end{enumerate}

We evaluate all tasks using binary task completion except for Task 1.
For Task 1, we adopt a more fine-grained evaluation inspired by target shooting: the innermost ring yields the highest score (5), while outer rings correspond to decreasing accuracy (4 to 1).
A score of 0 is assigned if the robot fails to grasp the object.
This scoring formulation captures subtle differences in spatial precision that would be lost in a binary metric.
We report the overall success rate (or average score in the case of Task 1) for each task. 
The object positions are marked by stickers to ensure fair comparison and reproducibility.

\noindent \textbf{Implementation Details.} For each task, we collect 100 expert demonstrations using tele-operation. To promote diversity and robustness, the positions of objects and targets are randomly perturbed during data collection.
In our real-world experiments, we keep the training and parameter settings identical to those used in simulation experiments.

\begin{table}[t]
\centering
\resizebox{\linewidth}{!}{
\begin{tabular}{lcccccc}
\toprule
 & Task1 (15) & Task2 (15) & Task3 (15) & Task4 (10) & Task5 (20)  & Average (75) \\ \midrule
$\pi_0$                    &    59.33\%            & 20.00\%              &  13.30\%           &  20.00\%           &  30.00\%    & 28.53\%      \\
Ours                          &   68.67\%             &  66.67\%             &   26.70\%          &     60.00\%       & 65.00\%     &57.41\%  \\ \bottomrule
\end{tabular}
}
\caption{\textbf{Success rates for five real-world tasks.} The number in parentheses represents the number of trials for each task. In particular, Task 1 is evaluated using scores; for example, 68.67\% denotes that our model achieves an average score of 3.43 out of a maximum of 5.}
\label{tab:real_com}
\end{table}

\noindent \textbf{Quantitative Results.}
We evaluate the effectiveness of our proposed method by comparing it against the baseline model $\pi_0$.
The quantitative results are presented in the Table~\ref{tab:real_com}.
Across all tasks, our method achieves consistent improvements over the baseline, indicating that the implicit 3D geometry features contribute positively to task performance.
Notably, our model demonstrates the largest performance gain on Task 2 (peg-in-hole insertion), a particularly challenging task that demands accurate spatial reasoning.
Furthermore, Task 3 (middle bottle grasping) poses a substantial challenge due to the narrow margin between adjacent bottles, requiring the gripper to perform careful, collision-free insertion and grasping.
Our method exhibits reasonable improvement on this task compared to the baseline, demonstrating enhanced spatial understanding and control in cluttered environments.
Overall, we achieve a 28.88\% performance gain in the average success rate.

\begin{figure}[t]
  \centering
  \includegraphics[width=\linewidth]{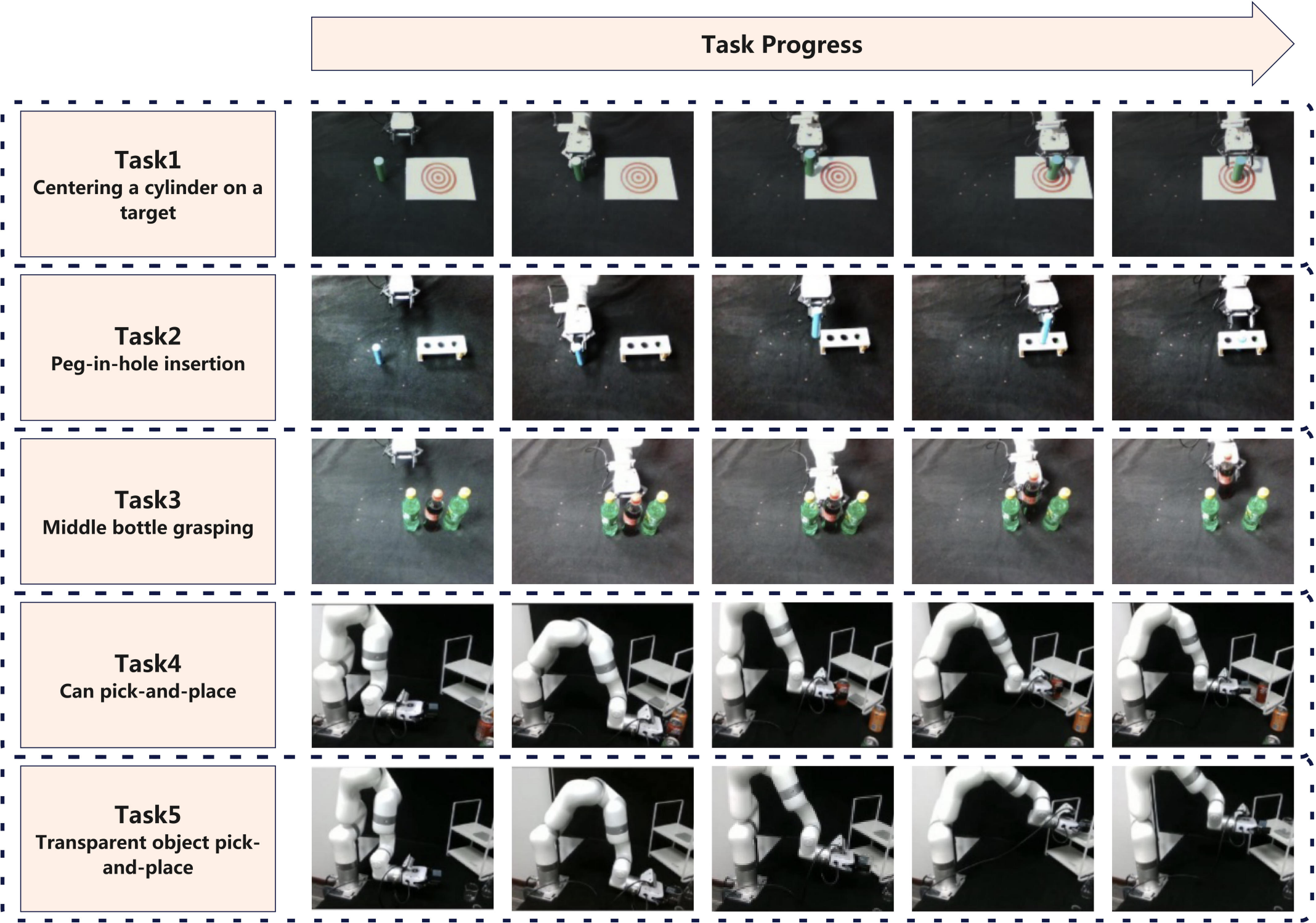}
  \caption{\textbf{Qualitative results in real-world tasks.} Detailed execution sequences of the five real-world evaluation tasks are visualized, illustrating the step-by-step progression for each scenario.}
  \label{fig:qua_results}
\end{figure}

\noindent \textbf{Qualitative Results.} 
In Figure~\ref{fig:qua_results}, we present visualizations of task executions across different tasks.
These visual results further complement the quantitative findings, showcasing our model's enhanced spatial awareness and manipulation precision.
For instance, in the cylinder-centering and peg-in-hole insertion tasks, our model reliably achieves stable grasping and precise alignment with the target area. 
In contrast, the baseline $\pi_0$ often fails to establish a proper grasp on the cylinder from the initial step, leading to unsuccessful or unstable placement attempts.

\subsection{Robustness Experiments}

\noindent \textbf{Task Setup.} 
In the robustness experiments, we begin with a simplified version of real-world Task~4 and progressively introduce additional disturbances to assess the model’s robustness. In this task, the robot is required to pick up a beverage can and place it onto a designated location on a shelf. Compared with the original Task~4 configuration, we apply two modifications to construct a base scenario: (i) the camera viewpoint is changed from a side view to a frontal view, mitigating visual occlusions during manipulation; and (ii) the shelf position is adjusted to facilitate easier placement, thereby reducing the likelihood of unintended collisions. Under this simplified condition, both $\pi_0$ and Evo-0 attain identical success rates. However, as disturbances are gradually introduced, Evo-0 consistently exhibits higher robustness than $\pi_0$.

\begin{figure}[t]
\vspace{3ex}
  \centering
  \includegraphics[width=0.98\linewidth]{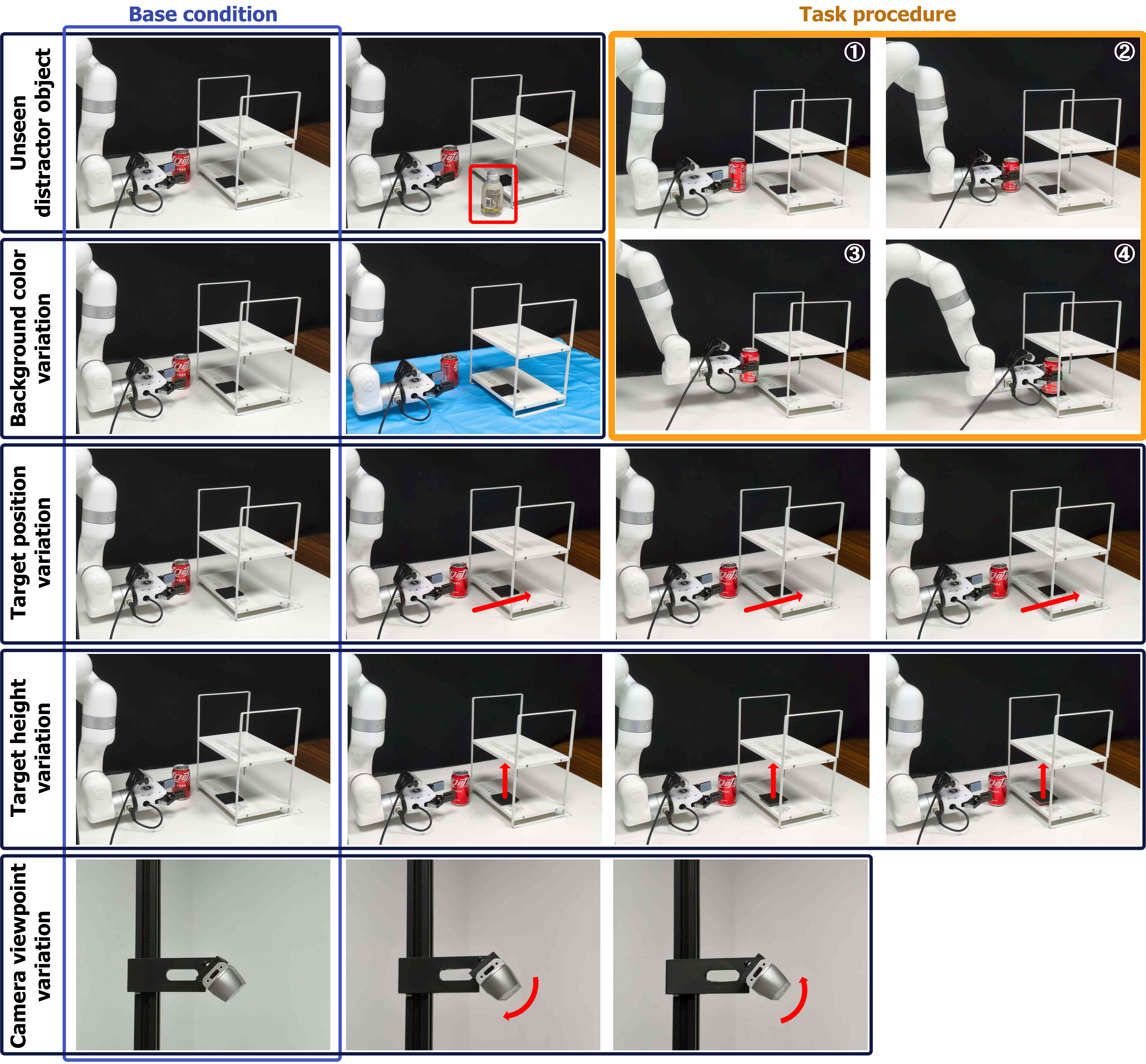}
  \caption{\textbf{Distractor settings of robustness experiments.} We visualize five disturbance conditions used in the robustness experiments: unseen distractor object, background color variation, target position variation, target height variation, and camera viewpoint variation. In the top right corner, we also present the execution sequence of a successful trial.}
  \label{fig:distractor}
\end{figure}

\noindent \textbf{Disturbance Conditions.} 
To systematically evaluate robustness, we design five categories of disturbance conditions, which are illustrated in Figure~\ref{fig:distractor}: (i) introduction of an unseen distractor object, (ii) variation in background color, (iii) displacement of the target position, (iv) change in target height, and (v) shift in camera viewpoint. All conditions lie outside the training distribution. Each disturbance targets a distinct aspect of generalization, enabling a comprehensive assessment of the model’s robustness to environmental variation and unseen distractors. 

\noindent \textbf{Quantitative Results.}  
We conduct comprehensive evaluations across all five disturbance settings, with the quantitative results summarized in Table~\ref{tab:disturbance_results}.

\begin{enumerate}
\item \textbf{Unseen distractor object.} 
In the unseen distractor object experiment, the target can and a distractor bottle are placed on opposite sides of the shelf. The task requires the robot to correctly identify the can, grasp it, and place it at the designated location. This setting evaluates both the model’s ability to distinguish the correct object and to execute precise placement in the presence of visual distractors. Experimental results show that Evo-0 substantially outperforms $\pi_0$ on both aspects. In the object selection phase, Evo-0 achieves a 100\% success rate, representing a 40 percentage point improvement over $\pi_0$ (60\%). For the full grasp-and-place sequence, Evo-0 attains a 70\% success rate, outperforming $\pi_0$ (20\%) by 50 percentage points. These results demonstrate that Evo-0 exhibits stronger robustness in both object discrimination and manipulation accuracy.

\item \textbf{Background color variation.} 
In the background color variation experiment, the tablecloth is changed to modify the scene appearance. This setting evaluates how each model maintains placement accuracy under background changes. Experimental results show that Evo-0 achieves approximately 10\% higher accuracy than $\pi_0$ in this condition. These findings indicate that Evo-0 possesses stronger robustness to variations in background appearance.

\item \textbf{Target position variation.} In the target position variation experiment, the mat is shifted 10\,mm, 20\,mm, and 30\,mm backward along the shelf across separate test groups. This setting assesses robustness to spatial displacement of the target. Evo-0 consistently achieves approximately 10\% higher success rates than $\pi_0$ under the 20\,mm and 30\,mm shifts, indicating stronger robustness and generalization to positional perturbations.

\item \textbf{Target height variation.} To assess robustness to vertical displacement, the thickness of the target mat is increased by 5\,mm, 10\,mm, and 15\,mm across separate test groups. As height increases, both models exhibit performance degradation. However, Evo-0 consistently outperforms $\pi_0$, with accuracy margins of 20\% and 10\% under the 10\,mm and 15\,mm conditions, respectively. These results indicate that Evo-0 demonstrates stronger robustness and enhanced spatial reasoning under vertical perturbations.

\item \textbf{Camera viewpoint variation.} 
In this setting, the wrist camera remains fixed, while the base camera is tilted upward and downward by 10\textdegree{} in separate experiments to introduce novel camera viewpoints. This variation evaluates robustness to changes in camera perspective. Both models experience performance degradation under these shifts; however, Evo-0 consistently outperforms $\pi_0$, achieving 20\% and 10\% higher accuracy in the upward and downward conditions, respectively. These results suggest that Evo-0 demonstrates stronger robustness to camera viewpoint changes.

\end{enumerate}

\begin{table}[]
\vspace{3ex}
\centering
\resizebox{0.90\linewidth}{!}{ 
\begin{tabular}{llrr}
\toprule
\multicolumn{2}{l}{\textbf{Condition}}                & \textbf{$\pi_0$} & \textbf{Ours}  \\ \hline 

\multicolumn{2}{l}{\textbf{Base}}                     & 70\%         & 70\%           \rule{0pt}{10pt} \\  \hline

\textbf{Unseen distractor object} &                      &              & \textbf{}      \rule{0pt}{10pt} \\
                               & Pick correct rate    & 60\%         & \textbf{100\%} \\
\textbf{}                      & Overall correct rate & 20\%         & \textbf{70\%}  \\ \hline
\textbf{Background color variation}     &                      &              & \textbf{}      \rule{0pt}{10pt} \\
                               & Add blue tablecloth  & 60\%         & \textbf{70\%}  \\ \hline
\textbf{Target position variation}    &                      &              &                \rule{0pt}{10pt} \\
                               & 10 mm backward       & 60\%         & 60\%           \\
                               & 20 mm backward       & 20\%         & \textbf{30\%}  \\
                               & 30 mm backward       & 10\%         & \textbf{20\%}  \\ \hline
\textbf{Target height variation}    &                      &              &                \rule{0pt}{10pt} \\
                               &5 mm higher         & 70\%         & 70\%           \\
                               & 10 mm higher         & 10\%         & \textbf{30\%}  \\
                               & 15 mm higher         & 0\%          & \textbf{10\%}  \\ \hline
\textbf{Camera viewpoint variation}  &                      &              &                \rule{0pt}{10pt} \\
                               & 10 degrees up        & 40\%         & \textbf{60\%}  \\
                               & 10 degrees down      & 30\%         & \textbf{40\%}  \\ \bottomrule
\end{tabular}
}
\caption{\textbf{Success rates for robustness experiments.} Comparison of success rates between $\pi_0$ and Ours under different disturbance conditions in real-world task robustness experiments.}
\label{tab:disturbance_results}
\end{table}

\section{Conclusion}
In this paper, we explore using implicit 3D representations to enhance spatial understanding in Vision-Language-Action (VLA) models. By leveraging features from the Visual Geometry Grounded Transformer (VGGT), trained on large-scale 2D–3D paired data, we inject strong geometric priors into VLA models without relying on explicit 3D inputs. Through extensive experiments across spatially challenging tasks in both simulation and real-world, we demonstrate that our approach significantly outperforms baseline models, validating the effectiveness of the proposed implicit geometric prior integration. Our method offers a simple and efficient solution for enhancing spatial understanding in VLA systems.

\bibliography{icra2026_conference}
\bibliographystyle{unsrt}


\end{document}